\documentclass[letterpaper, 10 pt, conference]{ieeeconf}

\IEEEoverridecommandlockouts
\usepackage{amsmath} 
\usepackage{amssymb}  
\usepackage{tensor}
\usepackage{mathtools}
\usepackage[binary-units=true]{siunitx}
\usepackage{booktabs}

\newcommand{\etalcite}[2]{#1~et~al.~\cite{#2}}

\title{\LARGE \bf Haptic Sequential Monte Carlo Localization \\
for
Quadrupedal Locomotion in Vision-Denied Scenarios}

\author{Russell Buchanan, Marco Camurri, Maurice Fallon
\thanks{The authors are with the Oxford Robotics Institute at the University of 
Oxford, United Kingdom. %
\texttt{\{russell, mcamurri, mfallon\}@robots.ox.ac.uk}}}

\usepackage{amssymb} 
\usepackage{algorithm2e} 
\usepackage{multirow} 

\usepackage{tikz}
\usetikzlibrary{bayesnet}
\usetikzlibrary{arrows,positioning}

\newcommand{\hide}[1]{}

\newcommand{\ie}{{i.e.,~}}

\newcommand{\bdmath}{\begin{dmath}}
\newcommand{\edmath}{\end{dmath}}
\newcommand{\beq}{\begin{equation}}
\newcommand{\eeq}{\end{equation}}
\newcommand{\bdm}{\begin{displaymath}}
\newcommand{\edm}{\end{displaymath}}
\newcommand{\bea}{\begin{eqnarray}}
\newcommand{\eea}{\end{eqnarray}}
\newcommand{\beal}{\beq \begin{array}{ll}}
\newcommand{\eeal}{\end{array} \eeq}
\newcommand{\beas}{\begin{eqnarray*}}
\newcommand{\eeas}{\end{eqnarray*}}
\newcommand{\ba}{\begin{array}}
\newcommand{\ea}{\end{array}}
\newcommand{\bit}{\begin{itemize}}
\newcommand{\eit}{\end{itemize}}
\newcommand{\ben}{\begin{enumerate}}
\newcommand{\een}{\end{enumerate}}

\newcommand{\Real}{\mathbb{R}}

\newcommand{\World}{\mathtt{W}}

\newcommand{\Base}{\mathtt{{B}}}

\newcommand{\meters}{\rm{m}}

\makeatletter
\let\NAT@parse\undefined
\makeatother
\usepackage{hyperref}
\begin{document}

\maketitle
\thispagestyle{empty} 
\pagestyle{empty}     

\graphicspath{{figures/}{../figures/}}
\hyphenation{proprio-ceptive}

\begin{abstract}
Continuous robot operation in extreme scenarios such as underground mines or 
sewers is difficult because exteroceptive sensors may fail due to fog, darkness,
 dirt or malfunction. So as to enable autonomous navigation in these kinds of 
situations, we have developed a type of proprioceptive
localization which exploits the foot contacts made by a quadruped robot to
localize against a prior map of an environment, without the help of any camera
or LIDAR sensor. The proposed method enables the robot to accurately re-localize
 itself after making a
sequence of contact events over a terrain feature. The method is
based on Sequential Monte Carlo and can support both 2.5D and 3D prior map
representations. We have tested the approach online and onboard the ANYmal
quadruped robot in two different scenarios: the traversal of a custom built
wooden terrain course and a wall probing and following task. In both scenarios,
the robot is able to effectively achieve a localization match and to execute a
desired pre-planned path. The method keeps the localization error down to
\SI{10}{\centi\meter} on feature rich terrain by only using its feet, kinematic
and inertial sensing.
\end{abstract}

\section{Introduction}
\label{sec:intro}
\begin{figure}
 \centering
 \includegraphics[width=0.85\columnwidth]{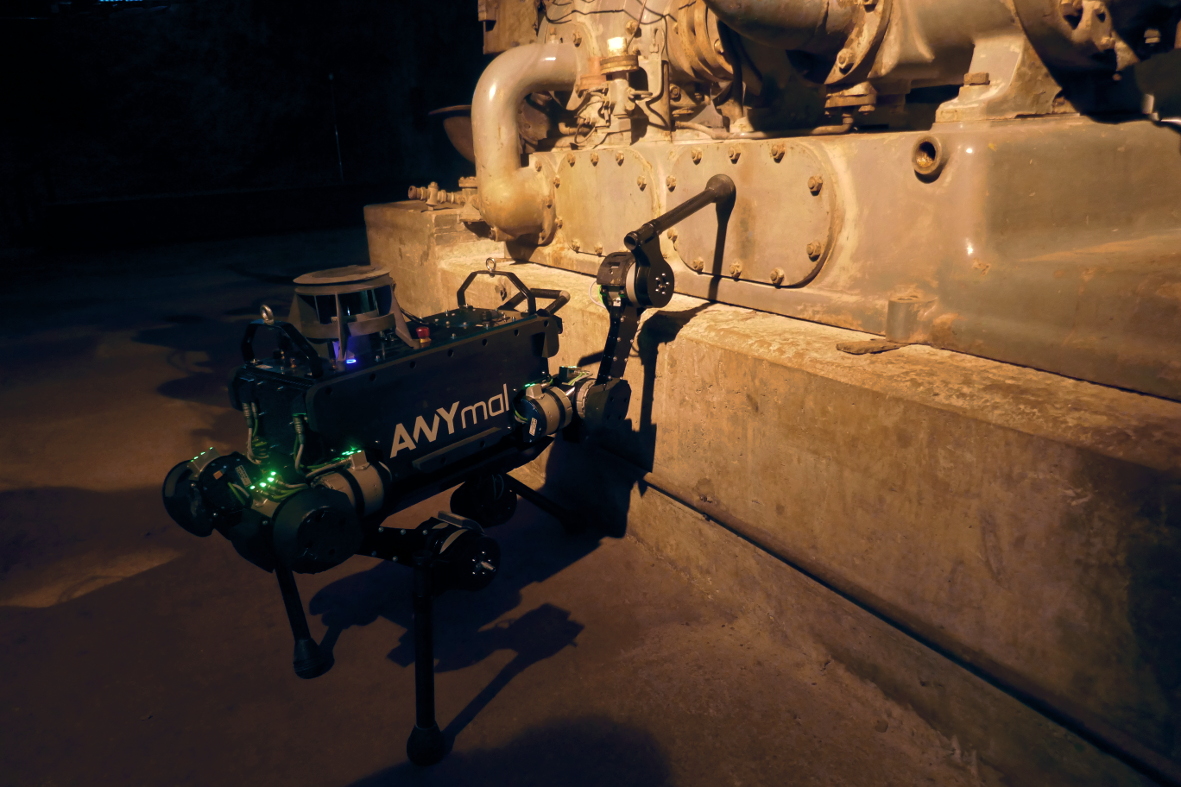}
 \caption{ANYmal using its foot to measure the vibrations of a piece of mining 
equipment. Tasks such as this might require localization even when vision cannot be used. By
exploiting prior knowledge of the 3D environment, the robot could use contact
events to localize itself. Source: ETH Zurich.}
\label{fig:sewer}
\end{figure}

Perception is an essential prerequisite for legged robot navigation on
challenging terrain. Several works have demonstrated the potential of such
machines when they are endowed with effective state
estimation~\cite{bloesch2017ral, nobili2017rss} and visual perception
capabilities~\cite{fankhauser2018icra,focchi2018heuristic,fallon2015humanoids}.
However, vision-denied areas such as dusty underground mines or foggy sewers are
places where legged robots are likely to be deployed \cite{Kolvenbach2020}
(see Fig. \ref{fig:sewer}). Additionally, unrecoverable sensor failures are
to be expected
when a robot is deployed for long-term autonomous missions. Performance should
degrade gracefully in such conditions by taking advantage of all proprioceptive sensor measurements and prior information available to the robot.

In this work, we explore haptics as the main source of robot
localization information on non-degenerate (i.e., non flat) ground. We present an algorithm for 6 Degrees of Freedom (DoF) localization of a quadrupedal robot based on Sequential Monte Carlo methods.

Given the pose (estimated by a proprioceptive state estimator
\cite{bloesch2017ral}), the joint kinematics and the foot contact states,
the robot is able to identify its most likely location within a prior map and 
its past trajectory. This is done using only proprioceptive sensors, such as Inertial Measurement Unit (IMU), encoders and torque sensors. To this end, the robot uses its legs to probe the terrain and uses the contact events
to estimate its trajectory relative to the terrain map. Unlike previous works, our method can localize in 6-DoF which we demonstrate with an experiment using a foot to probe in 3-dimensions. This problem is
analogous to a blindfolded person in a room they know well, using their hands to
localize within the room through touch.

We tested our algorithm using an ANYbotics ANYmal quadruped robot in two
different
scenarios. First, we constructed a diverse terrain course with features such as
ramps, chevrons and steps. The robot traversed several loops of this course, comparing
contact points to the prior map (represented as a 2.5D digital elevation map). In the
second scenario, the robot probed against two perpendicular walls to improve
its estimated pose, with the map represented as a 3D point cloud.

The remainder of this document is structured as follows:
Section~\ref{sec:related} summarizes the recent developments in the field of
legged
haptic localization and the related problem of in-hand tactile localization;
Section~\ref{sec:problem} defines the mathematical background of the legged
haptic localization problem; Section~\ref{sec:proposed} describes our proposed
haptic localization algorithm;
Section~\ref{sec:experimental} presents the experimental results collected using the
ANYmal
robot; Section~\ref{sec:discussion}
provides an interpretation of the results and discusses the limitations of the
approach; finally, Section~\ref{sec:conclusion} concludes the paper with final
remarks.

\section{Related Works}
\label{sec:related}
Pioneering work on how to use robot's legs not just for locomotion but also to
infer terrain information such as friction, stiffness and geometry has
been presented by Krotkov \cite{krotkov1990active}. This is a useful capability
when legged robots have to operate in vision-denied areas. The idea
was recently revisited by \etalcite{Bednarek}{bednarek2019icra} to classify
different terrain types in order to improve locomotion parameters and by
\etalcite{Kolvenbach}{Kolvenbach2020} to inspect
concrete in sewage systems by vibration analysis of the sewer floor.

While these works were more focused on using haptic information to improve
locomotion or to monitor the terrain, little research has been conducted
on how to use haptic information to localize the robot in the environment.
However, a similar problem has been explored in grasping: tactile localization.

Tactile localization involves the estimation of the 6-DoF
pose of an object (of known shape) in the robot's base frame by means
of kinematics of the robot's fingers and its tactile sensors. The pose of
the object (in robot base coordinates) is inferred by matching the set of
all positions of the sensed contacts (also in base coordinates) with the 3D
model of the object being grasped.

Intuitively, this problem is analogous to legged robot haptic localization, if
we imagine the entire robot as a hand trying to ``grasp'' the ground. The
objective of legged haptic localization is estimating the pose of the robot
relative to a fixed object (in this case the terrain) by means of its
``fingers'' (in this case the robot's feet), instead of the pose of the object
being grasped relative to the robot.

\subsection{Sequential Monte Carlo Methods}
Tactile localization has typically been addressed using Sequential Monte Carlo
(SMC)
methods,
a subfamily of which is called \emph{particle filters} \cite{Fox2001}. Since
the object
can have any shape, the probability distribution of its pose given tactile
measurements can be multimodal. This excludes all Gaussian estimators such as
Kalman filters. In contrast, SMC methods can maintain a
discrete approximation of an arbitrary distribution by generating many
hypotheses (also called particles) from a known \emph{proposal distribution}.
If the number of particles is large enough and the proposal distribution covers
the state space well (i.e., it captures the areas of high density of the true
distribution), the target distribution is well approximated by a list of state
particles and their associated importance weights.

SMC is sensitive to the dimension of the state space, which should be low enough to
avoid combinatorial explosions or particle depletion.
State-of-the-art methods aim to reduce this dimensionality and also to
sample the state space in an efficient manner. In the SLAM community, this has
been typically achieved by careful design of the proposal distribution and
adaptive importance resampling to avoid particle depletion
\cite{grisetti2005icra}.

\subsection{Tactile Localization in Manipulation}
Vezzani et~al. \cite{vezzani2017tro} proposed an algorithm for tactile localization  
using the Unscented
Particle Filter (UPF) \cite{vandermerwe2000nips} and tested it on the iCub robot
(equipped with contact sensors at the fingertips) to localize four different
objects in the robot's reference frame. The algorithm is recursive and can
process data in real-time as new measurements become available. The object
and the robot's base are assumed to be static, therefore the pose to be
estimated is a fixed parameter. This assumption allows the use of a window of
past measurements to better address the sparsity of the measurements, which
consist of a series of single finger touches. For
legged haptic localization, the assumption of both a static robot and terrain
does not hold in general and more general methods are required.

Chalon et~al. \cite{chalon2013online} proposed a particle filtering method for
online in-hand object localization that tracks the pose of an object while it
is being manipulated by a fixed base arm. The pose is subsequently used to
improve the performance of pick and place tasks (e.g., by placing a bottle in a
upright position). The particles representing the object's pose are initialized
by a Gaussian distribution around the true initial pose of the object, acquired
by a
vision system (used only at start). The particle weights are taken from
the measurements according to a function which penalizes both finger/object
co-penetration and the distance between the object and the fingertip that
detected the contact. The best estimate is the one with highest weight.

This work, as with many others in the field of grasping, assumes the hand can
envelop the object, since they have comparable sizes. This is not the case with
legged robots touching the ground, which makes the problem generally harder and
involves more uncertainty. For example, a hand touching a very large object
has fewer clues as to the object's pose, because the contact points would not
be equally spread across the object but concentrated in a small area of it.

\subsection{Haptic Localization of Walking Robots}
To the best our knowledge,~Chitta et~al.~\cite{chitta2007icra} is the only significant prior example of legged haptic localization to date.
In their work, they presented a proprioceptive localization
algorithm based on a particle filter for the LittleDog, a small electric quadruped.
The robot was commanded to perform a
statically stable gait over a known irregular terrain course, using a motion
capture system to feed the controller. While walking, the algorithm approximated
the probability
distribution of the base state with a set of particles. The state was reduced
from six to three dimensions: linear position on the $xy$-plane and yaw. Each
particle was sampled from the uncertainty of the odometry, while the weight of
a particle was determined by the L2 norm of the height error between the map
and the contact location of the feet. The algorithm was run offline on eight
logged trials of \SI{50}{\second} each.

In our proposed work, we improve upon \cite{chitta2007icra} both theoretically
and experimentally. First, we estimate the full past trajectory at every step
instead of the most recent pose. This allows us to estimate poses which are
globally consistent over the entire trajectory of the map. Second, we perform
localization in the full 6-DoF of the robot, instead of limiting the estimation
to the x, y and yaw dimensions. Third, we experimentally demonstrate the
effectiveness of the algorithm by running it online on our quadruped robot. This
is the first time such an algorithm has been successfully executed online on real
hardware and used in a closed loop navigation system to successfully execute a
planned path. Finally, we demonstrate the algorithm in three dimensions in online experiments.

\section{Problem Statement}
\label{sec:problem}

Let $\boldsymbol{x}_k = \left[ \mathbf{p}_k, \boldsymbol{\theta}_k\right]$ be
a robot's pose at time $k$, composed of the base
position in the world frame $\mathbf{p} \in \mathbb{R}^3$ and orientation
$\boldsymbol{\theta} \in
SO(3)$. With a slight abuse of notation, we will use the same symbol for its
homogeneous matrix form $\boldsymbol{x}_k \in SE(3)$.
We assume that for each timestep $k$, an estimate of the robot pose
$\tilde{\boldsymbol{x}}_k$ and its covariance $\Sigma_k \in
\mathbb{R}^{6\times6}$ is available from an odometric
estimator\footnote{uncertainties
for the rotation manifold are maintained in the Lie tangent space, as in
\cite{forster2017tro}}. We also assume that the location of the robot's end
effectors in
the base frame
($\mathbf{d}_\text{LF},\;\mathbf{d}_\text{RF},\;\mathbf{d}_\text{LH},\;\mathbf{d
}
_\text{RH}) \in \mathbb{R}^{3 \times 4}$ are known from forward
kinematics and their binary contact states $C \in
\mathbb{B}^4$ from inverse dynamics.

For simplicity,
we neglect any uncertainties due to inaccuracy in joint encoder readings or limb
flexibility. Therefore,
the propagation of the uncertainty from the base to the end effectors is
straightforward to compute. For brevity, the union of the aforementioned states
(pose and contacts)  at time $k$ will be referred as the \emph{quadruped state}
$\mathcal{Q}_k = \{
\tilde{\mathbf{p}}_k, \tilde{\boldsymbol{\theta}}_k, \Sigma_k,
\mathbf{d}_\text{LF},\;\mathbf{d}_\text{RF},\;\mathbf{d}_\text{LH},\;\mathbf{d}
_\text{RH},\; C_k\}$.

Our approach can localize against both 2.5D terrain elevation
maps and 3D point clouds. These prior maps
are denoted with $\mathcal{M}$.

Our goal is to use a sequence of quadruped states, and their corresponding
uncertainties
$\Sigma_k$, to estimate the most likely sequence of robot states up to time $k$
as:
\begin{equation}
\mathcal{X}^{*}_{0:k} =
[\boldsymbol{x}^{*}_0,\;\boldsymbol{x}^{*}_1,\;\dots,\;\boldsymbol{x}^{*}_k]
\end{equation}
such that the likelihood of the contact points to be on
the map is maximized.

\section{Proposed Method}
\label{sec:proposed}
To perform localization, we sample a predefined number of particles at
regular intervals from the pose distribution provided by the odometry, as
described in Section~\ref{sec:sampling}, and we compute the likelihoods of
the measurements by comparing each particle to the prior
map so as to update the weight of the particle filter (see
Section~\ref{sec:measurement}).
For convenience, we give a brief summary of particle filter theory in
Section~\ref{sec:primer}.

\subsection{Particle Filter Theory}
\label{sec:primer}
In a particle filtering framework, the objective is to approximate the
posterior distribution of the state $\boldsymbol{x}_k$
given a history of measurements $\boldsymbol{z}_0,\dots,\boldsymbol{z}_k
= \boldsymbol{z}_{0:k}$ as follows:

 \begin{equation}
 p\left(\boldsymbol{x}_k | \boldsymbol{z}_{0:k}\right) = \sum_i w^i_{k-1}
 p\left(\boldsymbol{z}_k |
\boldsymbol{x}_k\right)p\left(\boldsymbol{x}_k|\boldsymbol{x}^i_{k-1}\right)
 \end{equation}
where $w^i$ is the \emph{importance weight} of the $i$-th particle;
$p\left(\boldsymbol{z}_k |
\boldsymbol{x}_k\right)$ is the measurement likelihood function and
$p\left(\boldsymbol{x}_k|\boldsymbol{x}^i_{k-1}\right)$ is the dynamic
model for the $i$-th particle state. Since $p\left(\boldsymbol{x}_k |
\boldsymbol{z}_{0:k}\right)$ is normally unknown, the state $\boldsymbol{x}_k$
is typically sampled from
$p\left(\boldsymbol{x}_k|\boldsymbol{x}^i_{k-1}\right)$, yielding:
 \begin{equation}
 p\left(\boldsymbol{x}_k | \boldsymbol{z}_{0:k}\right) = \sum_i w^i_{k-1}
 p\left(\boldsymbol{z}_k |
\boldsymbol{x}^i_{k-1}
\right)\delta\left(\boldsymbol{x}_k - \boldsymbol{x}_k^i\right)
 \end{equation}
where $\delta(\cdot)$ is the Dirac delta function.

\subsection{Locomotion Control and Sampling Strategy}
\label{sec:sampling}
In the considered situation, the robot would be locomoting blindly. As a result,
it is likely to do so with a statically stable gait while using the terrain to
achieve a position fix.

Statically stable gaits alternate between three and
four points of support by moving a single foot. When a new four-point support
phase occurs at time $k$, the estimated robot state $\mathcal{Q}_k$ from the
onboard state estimator is collected.
Similarly to \cite{chitta2007icra}, particles are sampled from
the following proposal distribution:
\begin{equation}
 p(\boldsymbol{x}_k | \boldsymbol{x}_{k-1}^i) =
\mathcal{N}(\boldsymbol{x}_k, \Delta\tilde{\boldsymbol{x}}_{k}
\boldsymbol{x}_{k-1}^i, \Sigma_k)
\end{equation}
where $\Delta\tilde{\boldsymbol{x}}_{k} =
\tilde{\boldsymbol{x}}_{k-1}^{-1}\tilde{\boldsymbol{x}}_{k}$ is the relative
pose between the estimated states at times $k-1$ and $k$. 

The new
samples $\boldsymbol{x}_k^i$ are drawn from a Gaussian centered at the pose of
the previous iteration composed of the relative pose from the odometry, with
the covariance of the odometry at time $k$. Since the estimator
\cite{bloesch2017ral} fuses both IMU and leg odometry, the covariance $\Sigma_k$
has very low uncertainty on roll and pitch because these two quantities are
observable due to gravity. This would result in all particles having virtually the same roll and pitch which permits reduction of the sample space to the four
other dimensions. This allows us to localize in the full 6-DoF space without incurring
problems due to high dimensionality.

\subsection{2.5-Dimensional Measurement Likelihood Model}
\label{sec:measurement}
\begin{figure}
 \centering
 \vspace{5pt}
\includegraphics[width=0.350\columnwidth]{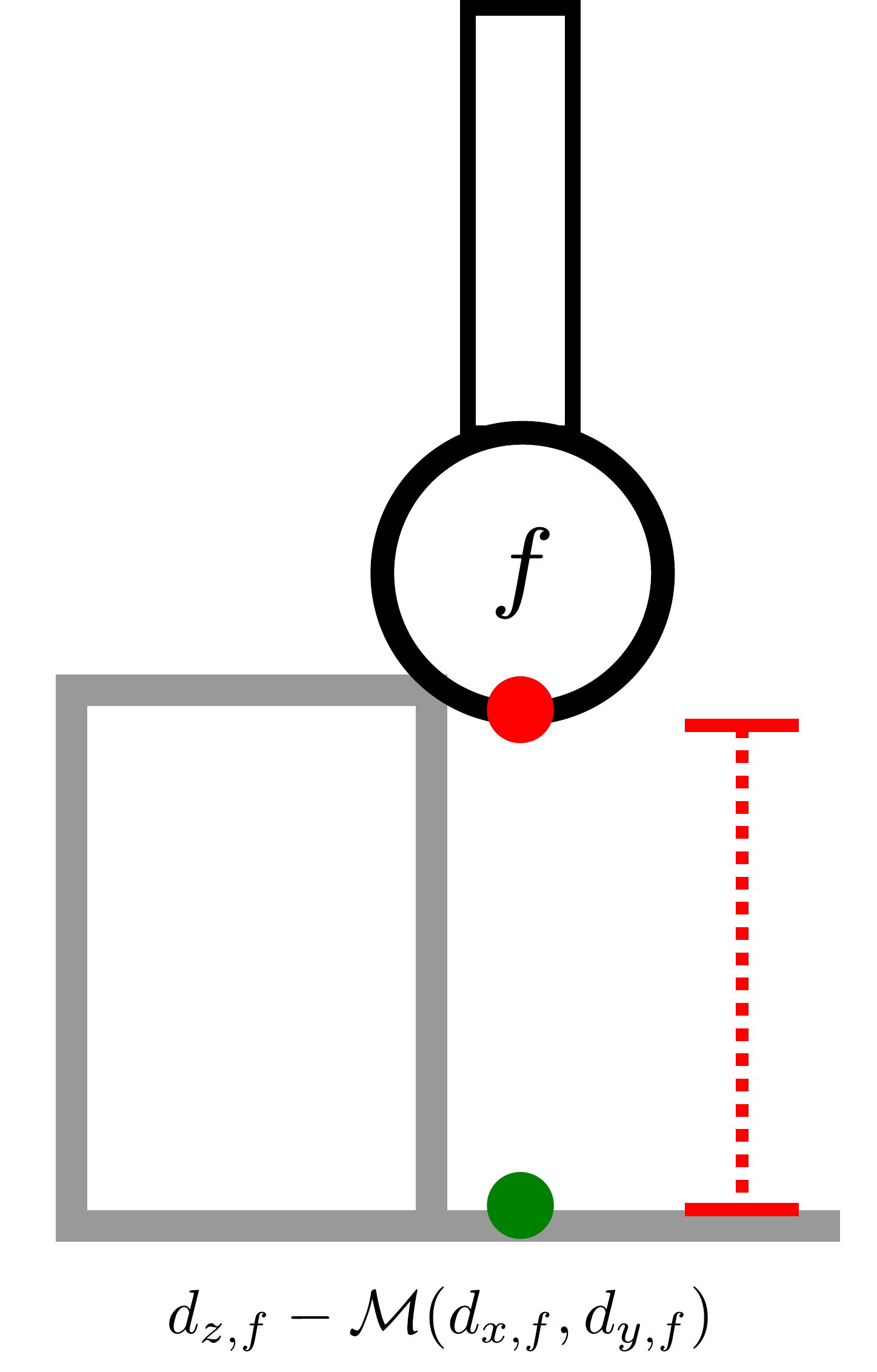}
\hspace{0.4in}
\includegraphics[width=0.350\columnwidth]{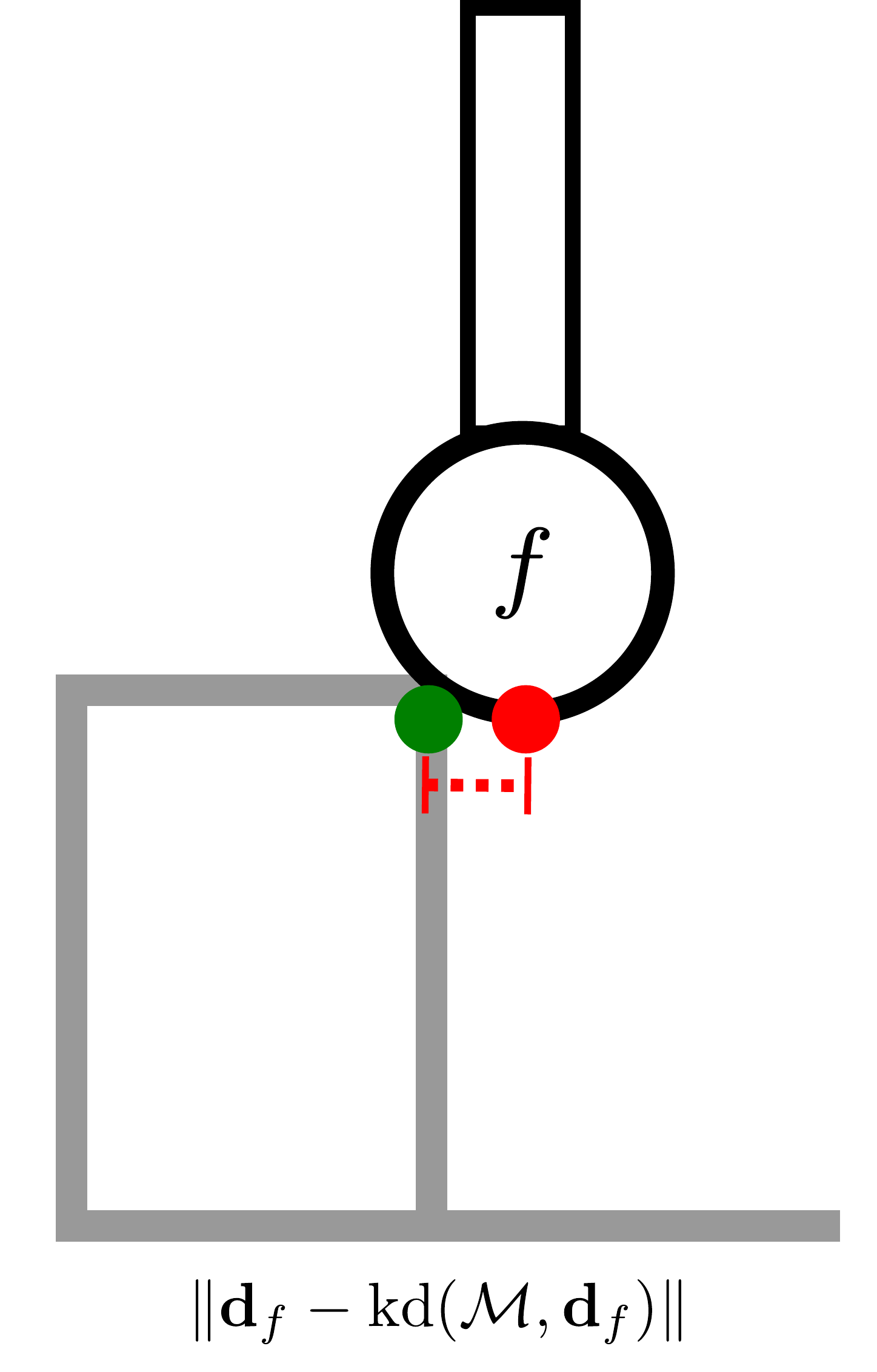}
 \caption{Comparison between contact measurements for 2.5D (left)
and 3D (right)
map representations. The red dots indicate the contact point as sensed by
the robot, while the green dots indicate the corresponding location returned
by the map. The red line shows the magnitude of the measurement. Note how
the same contact interaction generates a much lower likelihood in
the 2.5D map than the 3D map, which is more representative.
This is a practical limitation of using a 2.5D map.}
 \label{fig:error}
\end{figure}

To account for uncertainties in the prior map, we model the measurement
likelihood as a univariate Gaussian centered at the local elevation of each
cell, with a manually selected variance $\sigma_z$ (\SI{1}{\centi\meter} in our
case).
In contrast with \cite{chitta2007icra}, we evaluate each contact point
individually. Given a particle state $\boldsymbol{x}^i_k$, the
estimated position of a contact in world coordinates for foot $f$ is
defined as the concatenation of the estimated robot base pose
and the location of the end effector in base coordinates (Fig. \ref{fig:error}
left):
\begin{equation}
\mathbf{d}^i_f = (d^i_{xf}, d^i_{yf}, d^i_{zf},) =
\boldsymbol{x}^i_k \mathbf{d}_f
\end{equation}
Thus, the measurements and their relative likelihood functions for the
$i$-th particle and a specific foot $f$ are:
\begin{align}
 z_k &= d^i_{zf} - \mathcal{M}(d^i_{xf},d^i_{yf})\\
p(z_k | \boldsymbol{x}^i_k) &=
\mathcal{N}(z_k,0,\sigma_z)
\label{eq:2dlikelihood}
\end{align}
where: $d^i_{zf}$ is the vertical component of the estimated contact
point location in world coordinates of foot $f$,  according to the $i$-th
particle; $\mathcal{M}(d^i_{x,f},d^i_{y,f})$ is the
corresponding map elevation at the  $xy$ coordinates of
$\mathbf{d}_f^i$.

\subsection{3-Dimensional Measurement Likelihood Model}
\label{sec:measurement3d}
Our method can incorporate contact events from 3D probing. This is useful for areas where the floor does not provide enough information to localize. Instead, the robot can probe walls and 3D objects.
In this case, we instead represent the prior map $\mathcal{M} \in
\Real^{3 \times N}$ by a 3D point cloud with $N$ points.
The likelihood of a particular contact point is computed using the Euclidean
distance
between the foot and the nearest point in the map.
This likelihood is again modeled as a zero-mean Gaussian
evaluated at the Euclidean distance between the estimated contact point
$\mathbf{d}^i_{f}$ and its nearest neighbor on the map, with variance
$\sigma_z$:
\begin{align}
 z_k &= \lVert\mathbf{d}^i_{f}-\text{kd}(\mathcal{M},\mathbf{d}^i_{f})\rVert
\\
p(z_k | \boldsymbol{x}^i_k) &=
\mathcal{N}(z_k,0,\sigma_z)
\label{eq:3dlikelihood}
\end{align}
where $\text{kd}(\mathcal{M},\mathbf{d}^i_{f})$ is the function that returns
the nearest neighbor of $\mathbf{d}^i_{f}$ on the map $\mathcal{M}$,
computed from its k-d tree.

\section{Implementation}
Pseudocode for our method is listed in Algorithm~\ref{alg:pf}.
\begin{algorithm}
	\vspace{2pt}
$\boldsymbol{x}_0^i \sim \mathcal{N}(\boldsymbol{x}_0,
\tilde{\boldsymbol{x}}_{0}, \Sigma_0)\quad \forall i \in N$

\ForEach{four-support phase $k$}{
	$\Delta\tilde{\boldsymbol{x}}_{k} \leftarrow%
	\tilde{\boldsymbol{x}}_{k-1}^{-1}\tilde{\boldsymbol{x}}_{k}$

	\ForEach{particle $i \in N$}{
		$\boldsymbol{x}^i_k \sim \mathcal{N}(\boldsymbol{x}_k,
		\Delta\tilde{\boldsymbol{x}}_{k}
		\boldsymbol{x}_{k-1}^i, \Sigma_k)$

		$w^i_k \leftarrow w^i_{k-1}$

		\ForEach{foot $f$} {
		\eIf{$\text{2.5D}$}{
		  $z_k \leftarrow d^i_{zf}$
		  }{
		  $z_k \leftarrow \lVert\text{kd}(\mathcal{M},\mathbf{d}^i_{f}) -%
\mathbf{d}^i_{f}\rVert$
          }
		  $w^i_k \leftarrow w^i_k p(z_k | \boldsymbol{x}^i_k)$
	    }

	$ x_k^* \leftarrow \text{WeightedMean}(x_k^0 \dots x_k^N, w_k^0,\dots
w_k^N)$

	$\mathcal{X}_k^* \leftarrow [\boldsymbol{x}_0^j,\dots, \boldsymbol{x}_k^j]$
	
	\If{$variance(w_k^i) > threshold$}
	{$resample(x_k^i)$}

}}
\caption{Haptic Sequential Monte Carlo Localization}
\label{alg:pf}
\end{algorithm}
We update the filter when the robot enters a new four-support configuration. At
this time $k$, the estimated robot pose
$\tilde{\boldsymbol{x}_k}$ is collected and used to compute the relative motion
$\Delta\tilde{\boldsymbol{x}}_{k-1:k}$. This is used to propagate forward the
state
of each particle $\boldsymbol{x}^i_k$ and draw a sample from the distribution
centered in $\Delta\tilde{\boldsymbol{x}}_{k}
\boldsymbol{x}_{k-1}^i$ with covariance $\Sigma_k =
(\sigma_{x,k},\sigma_{y,k},\sigma_{z,k})$.

The weight of a particle $w^i$ is then updated by multiplying it by the
likelihood that each foot is in contact with the map. In our implementation, we
modify the likelihood functions from Equations \ref{eq:2dlikelihood},
\ref{eq:3dlikelihood} as:
\begin{equation}
p(z_k | \boldsymbol{x}^i_k) = \min(\rho,\mathcal{N}(z_k,0,\sigma_z))
\end{equation}
where $\rho$ is a minimum weight
threshold, so that outlier contact measurements do not immediately lead to
degeneracy. Resampling is triggered when the variance of the weights rises
above a certain threshold. This was manually tuned once and used for all experiments. Resampling is necessary otherwise the particle set
would disperse across the state space and many particle would have low weight.
By triggering this process when the variance of the weights increases, the
particles can first track the dominant modes of of the underlying distribution.

\subsection{Particle Statistics}
\label{sec:statistics}
The optimal value for the $k$-th iteration,
$\boldsymbol{x}^*_k$, is computed from the weighted average of all the
particles, as in \cite{chitta2007icra}. However, this method does not take into
account the multi-modality of the particle distribution. There can be several
hypotheses resulting in multiple clusters of particles whose weighted average is
not a reliable estimate of the robot's location. To account for this, we
check the variance of the particle positions in the $x$ and $y$ axes. If they
are
low (\ie $\ll \sigma_{x,k},\sigma_{y,k}$), we assume a good estimate and
update the robot's full pose. However, if they are high, we update only the
$z$ component of the robot's location (the main drifting dimension). The other dimensions drift
at the same rate as the state estimator, but the upward drift is corrected
since the robot is always in contact with the ground.
\begin{figure}
	\vspace{4pt}
	\centering
	\includegraphics[width=\columnwidth]{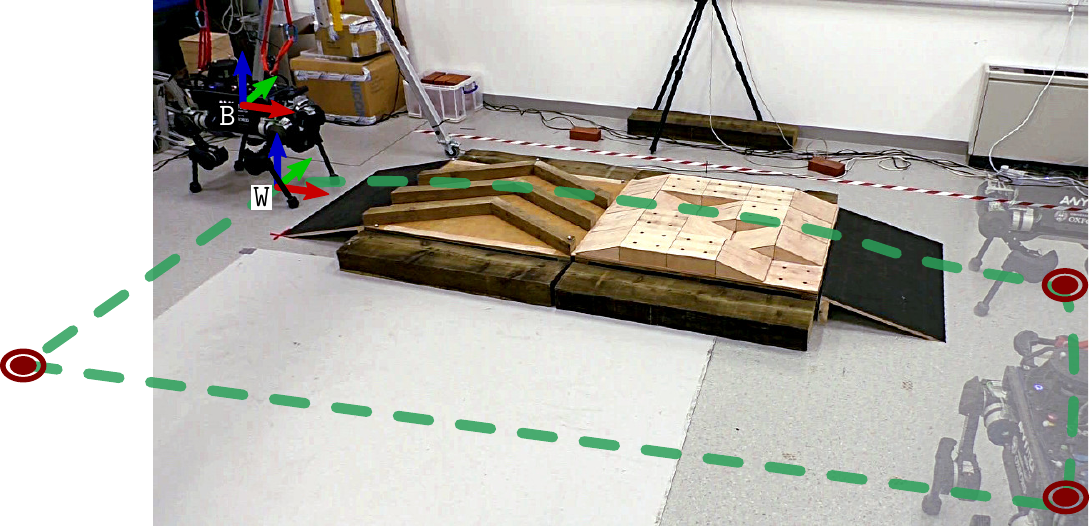}
	\caption[ANYmal haptic localization experiment]{ANYmal haptic
		localization experiments. The robot traverses the terrain, turns 90 deg right
		and comes back to the initial position passing trough the flat area. The goals
		given to the planner are marked by the dark red disks, while the planned
		route is a dashed
		green line (one goal is out of the camera field of view). The
		world
		frame
		$\World$ is fixed to the ground, while the base frame $\Base$ is rigidly
		attached to the robot's chassis. The mutual pose between the robot and the
		terrain course is bootstrapped with the Vicon.}
	\label{fig:anymal-chevron}
\end{figure}
\begin{figure}
\centering
\includegraphics[width=\columnwidth]{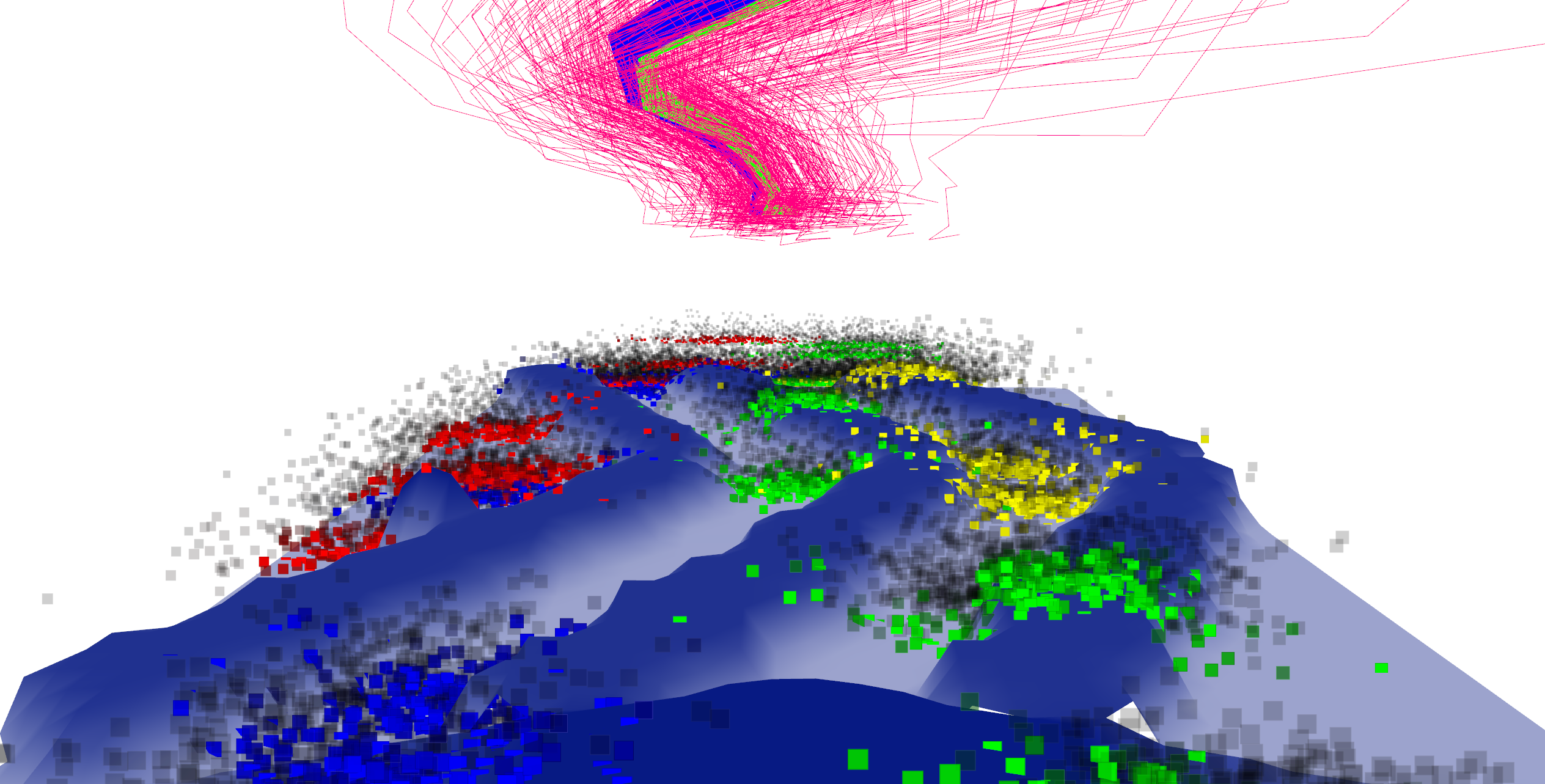}
\includegraphics[width=\columnwidth]{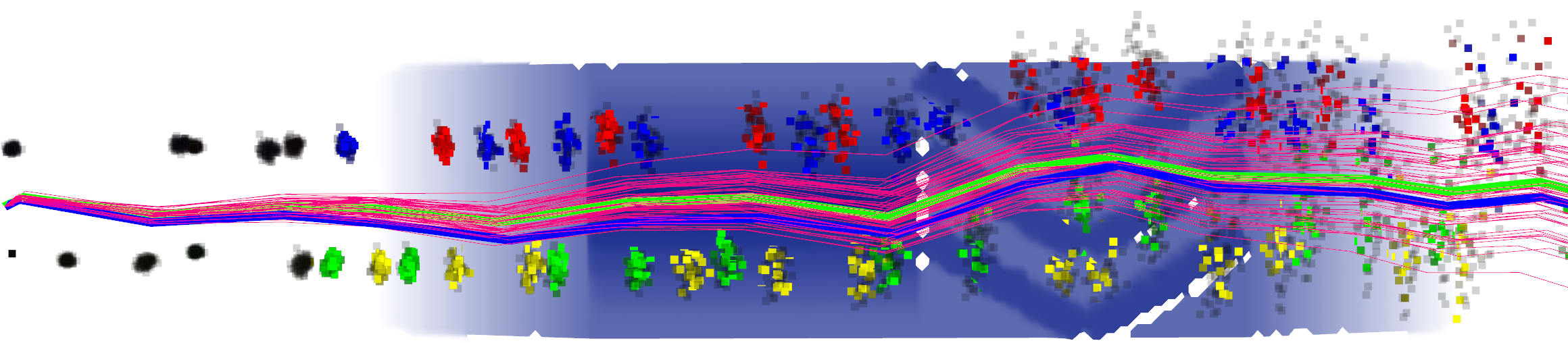}
 \caption{\emph{Top:} Perspective view of a simulated traversal of similar terrain to Fig.~\ref{fig:anymal-chevron}. At the top is the 1000 sampled
trajectories shown by thin magenta lines, while the ground truth and estimated trajectory are shown with a green and blue line, respectively. \emph{Bottom:} Top down view of the same scene. The top perspective is from the center the terrain facing right. Each squared dot represents one of 4000 contacts,
colored by leg type (LF: red, RF: green, LH: blue, RH: yellow). The tone and
alpha of each square is associated to the likelihood: darker and more
transparent relates to a lower likelihood. The grid map is colored by
height, from white to dark blue. }
 \label{fig:contact}
 \vspace{-10pt}
\end{figure}

\section{Experimental Results}
\label{sec:experimental}
Our localization algorithm has been tested in two different scenarios: a
walking task crossing a
custom built terrain course and a task involving probing and following of a
wall.
The prior maps were captured with a Leica BLK-360 laser
scanner, which provides registered point clouds with sub-centimeter accuracy.

Before testing on real hardware, the 3D model of the terrain course was
imported as a simulated environment (in Gazebo) allowing us to validate the
localization
algorithm with different values of noise and particle set dimensions. Fig.
\ref{fig:contact}
shows an example with 1000 particles depicted as candidate trajectories (in
magenta) with the associated contacts colored by their likelihood
(brighter/more solid color = higher likelihood). The most likely trajectory is
highlighted in blue and is the closest to the ground truth (in green).

In real world experiments, the ground truth robot poses were collected at
\SI{100}{\hertz} with millimetric accuracy using a
Vicon motion capture system. At start of the experiment, the relative position
of the robot within the
map was measured using Vicon and used for initialization only. Thereafter, the
pose of the robot was estimated
using the particle filter and the Vicon was used only to compute error
statistics. To account for initial errors, particles at the start were sampled
from
a Gaussian centered at the initial robot pose with a covariance of
\SI{20}{\centi\meter}.

\begin{table}
	\vspace{4pt}
\centering
\resizebox{\columnwidth}{!}{
\begin{tabular}{lccccc}
\toprule
Exp. & Dist. [\si{\meter}] & Time [\si{\second}] & Loops & ATE TSIF & ATE HL \\
\toprule
1 & 66.42 & 525 & 2 &0.63 & \textbf{0.13} \\
\midrule
2 & 145.31 & 1097 & 4 & 2.57 & \textbf{0.40}  \\
\midrule
3 & 55.67 & 557 & 2 & 0.52   & \textbf{0.19} \\
\midrule
4 & 68.71 & 604 & 2 & 0.65 & \textbf{0.32}  \\
\midrule
5 & 172.65 & 1606 & 4 & 2.00 & \textbf{0.61}  \\
\bottomrule
\end{tabular}}
 \caption{Estimation performance on the terrain course datasets. ATE  =
Absolute Translation Error; TSIF = Two-State Implicit Filter
\cite{bloesch2017ral} ; HL = Haptic
Localization. }
\label{tab:rpe}
\end{table}

\subsection{Terrain Course Experiments}
In the first scenario, the robot was commanded to navigate between four walking 
goals at the corners of a rectangle. One of the edges required crossing 
a \SI{4.2}{\meter} terrain course composed of a
\SI{12}{\degree} ascending ramp, a \SI{13}{\centi\meter} high chevron pattern,
an asymmetric composition of uneven square blocks and a
\SI{12}{\degree} descending ramp (Fig. \ref{fig:anymal-chevron}). After
crossing the wooden course, the robot returned to the starting
position across a portion of flat ground, which tests how the system behaves in
feature-deprived
conditions.

While impressive, blind reactive locomotion has been developed by a number of
groups
including \cite{dicarlo2018iros,focchi2018heuristic}, unfortunately our
haptic controller was not
sufficiently reliable to cross the terrain course so we resorted to use of the
statically stable gait from \cite{fankhauser2018icra}
which used a depth camera to aid footstep planning. Using our
approach, localization was performed
without access to any camera information.

To demonstrate repeatability, we have performed five experiments of this type,
for a total distance traveled
of more than \SI{0.5}{\kilo\meter} and \SI{1}{\hour}\SI{13}{\minute} duration.
A summary of the experiments is presented in
Table \ref{tab:rpe}, where the haptic localization shows an overall improvement
between \SIrange{50}{85}{\percent} in the Absolute Translation Error (ATE).
With the haptic localization, the ATE is \SI{33}{\centi\meters} on average,
which reduces to \SI{10}{\centi\meters} when evaluating only the
feature-rich portion of the experiments (\ie the terrain course traversal).

\begin{figure}
	\vspace{4pt}
 \centering
 \includegraphics[width=\columnwidth]{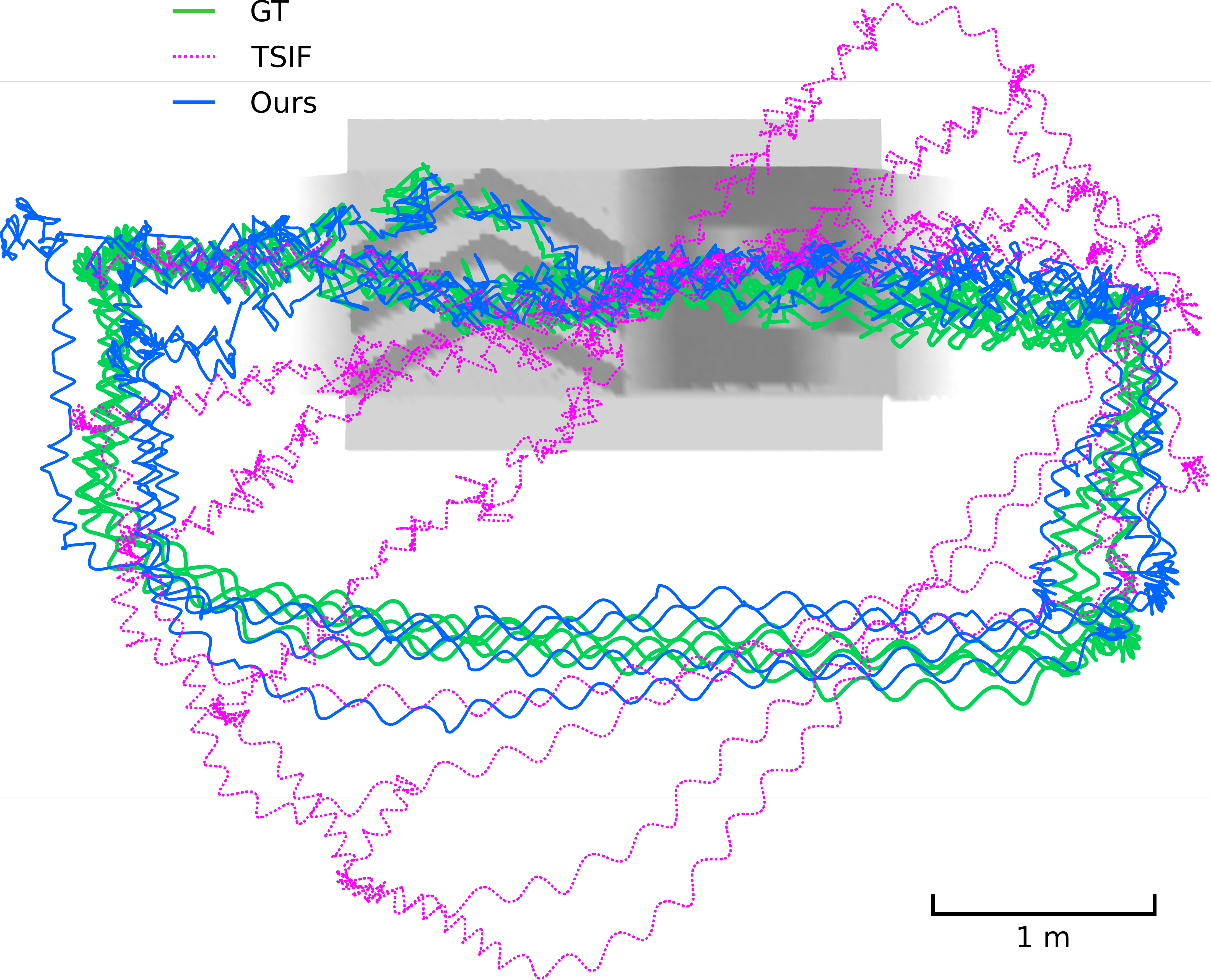}
 \caption{Top view of the estimated trajectories from TSIF (dashed magenta),
haptic localization (blue) and ground truth (green) for Experiment 2.}
 \label{fig:top-view}
 \vspace{-10pt}
\end{figure}

For Experiments 1 and 2, the robot was manually operated to traverse the terrain
course, completing two and four loops, respectively.
In Experiments 3--5, the robot was commanded to follow the rectangular path
autonomously. In these
experiments, the haptic localization algorithm was run online and in closed-loop
and effectively
guided the robot towards the goals (Fig. \ref{fig:top-view}).
Using only
the prior knowledge of the terrain geometry and the contact events, the robot
stayed localized
in all the runs and successfully tracked the planned goals. This can be seen in
Fig.
\ref{fig:top-view}, where the estimated
trajectory (in blue) drifts on the $xy$-plane when the robot is walking on the
flat ground, but it becomes re-aligned with the ground truth when the robot
crosses the
 terrain course again.

\begin{figure}
	\vspace{4pt}
 \centering
 \includegraphics[width=\columnwidth]{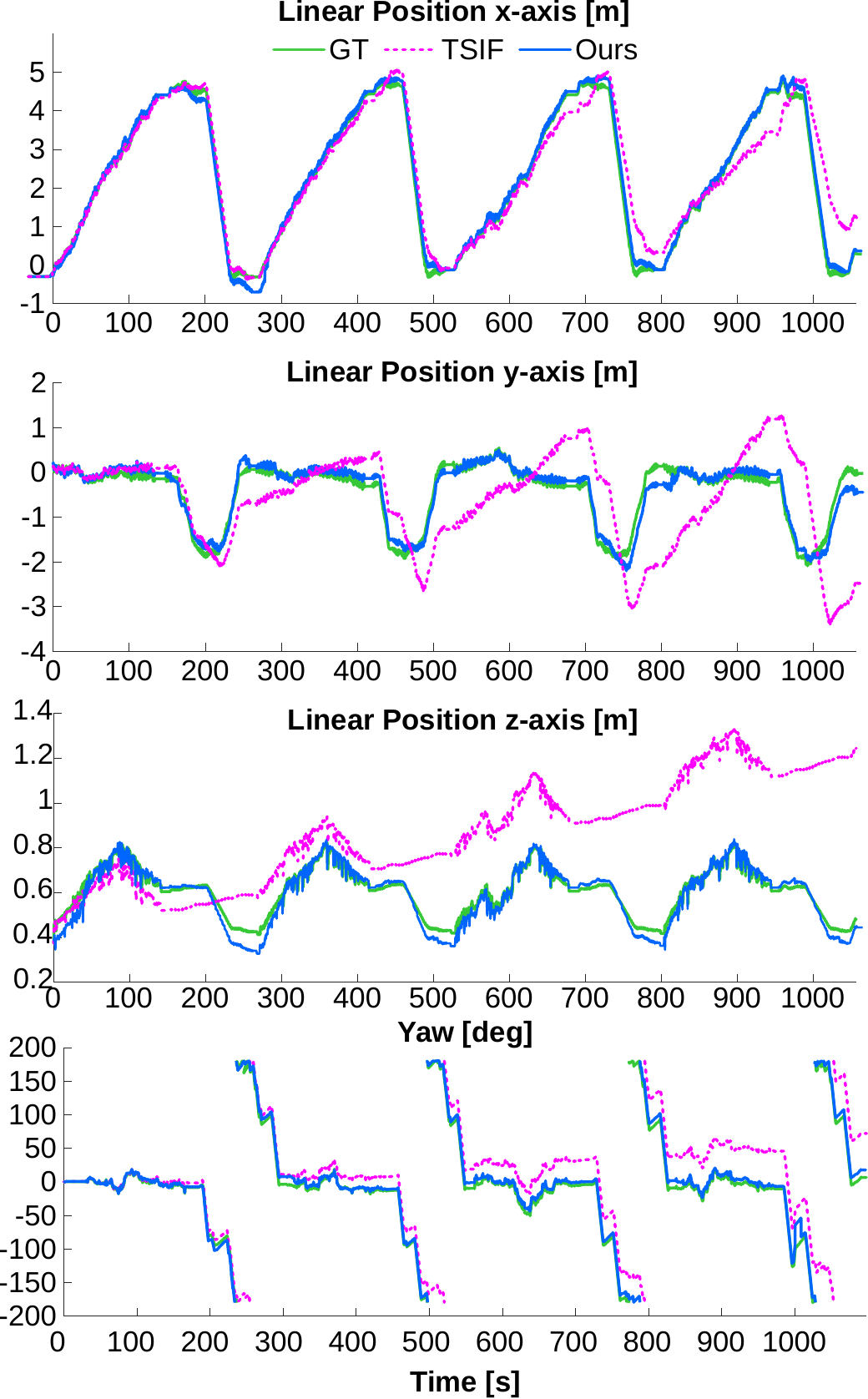}
 \caption{Comparison between the estimated position from TSIF (dashed magenta)
and haptic localization (blue) against Ground Truth (green) for Experiment 2.
After
\SI{200}{\second}, the estimation error in TSIF has drifted significantly upward
and in yaw.
 In particular, the upward drift is noticeable in the third plot, where
the values grow linearly. The drift is eliminated by the re-localization against
the prior map.}
 \label{fig:plots}
 \vspace{-10pt}
\end{figure}

Fig. \ref{fig:plots} shows in detail the estimator performance for each of
the three linear dimensions and yaw. Since position and yaw are unobservable,
the drift on these states is unbounded. In particular, the error on the odometry
filter
(TSIF \cite{
bloesch2017ral}, magenta dashed line) is dominated by upward drift (due to
kinematics errors and impact nonlinearities,
see third plot) and
yaw drift (due to IMU bias, see bottom plot). This drift is estimated and
compensated for by
the haptic localization (blue solid line), allowing accurate tracking of the
ground truth
(green solid line) in all dimensions. This can be
noted particularly at the four peaks in the $z$-axis plot, where the estimated
trajectory and
ground truth overlap. These times coincide with the robot's summit of the terrain course.

\subsection{Online Haptic Exploration on Vertical Surfaces}
The second scenario involved a haptic wall following task: the robot started in
front of a wall with an uncertain location. The particles were again initialized
with
\SI{20}{cm} position covariance. To test the capability to recover
from an initial error, we applied a \SI{10}{cm} offset in both $x$ and
$y$ from the robot's true position in the map. At start, the robot was
commanded to walk \SI{1}{m} to the right (negative $y$ direction) and press a button on the wall. To accomplish the task, the robot needed to ``feel its way'' by alternating probing motions with its right front foot and walking laterally to localize inside the room. The fixed number of probing motions was pre-scripted so with each step to the right, the robot probed both in front and to its right. In this scenario, the prior map is represented as a 3D point cloud with a minimum distance of \SI{1}{\centi\meter} between its points. The whole experiment was executed blindly with the static controller from \cite{fankhauser2018icra}.

\begin{figure}
	\vspace{5pt}
	\centering
	\includegraphics[width=0.240\columnwidth]{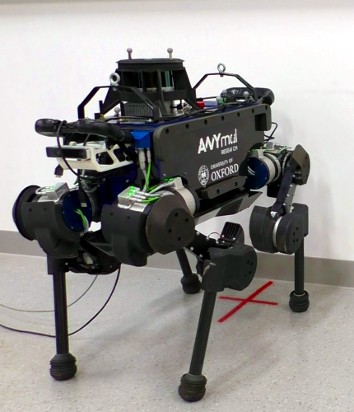}
	\includegraphics[width=0.240\columnwidth]{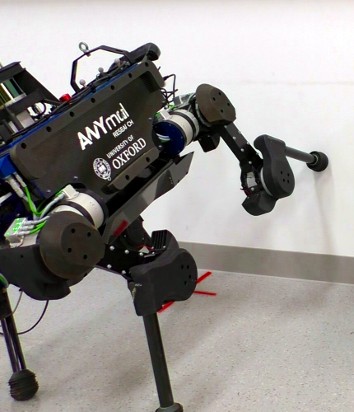}
	\includegraphics[width=0.240\columnwidth]{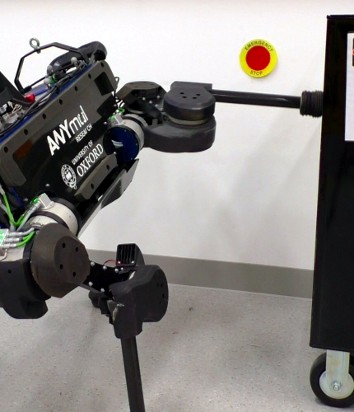}
	\includegraphics[width=0.240\columnwidth]{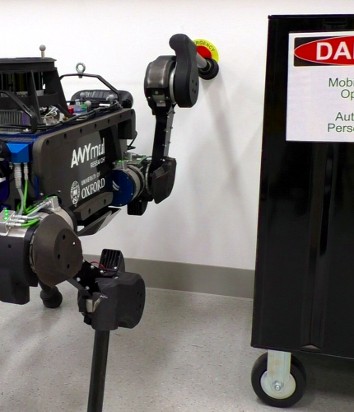}\\
	\vspace{2pt}
	\includegraphics[width=0.240\columnwidth]{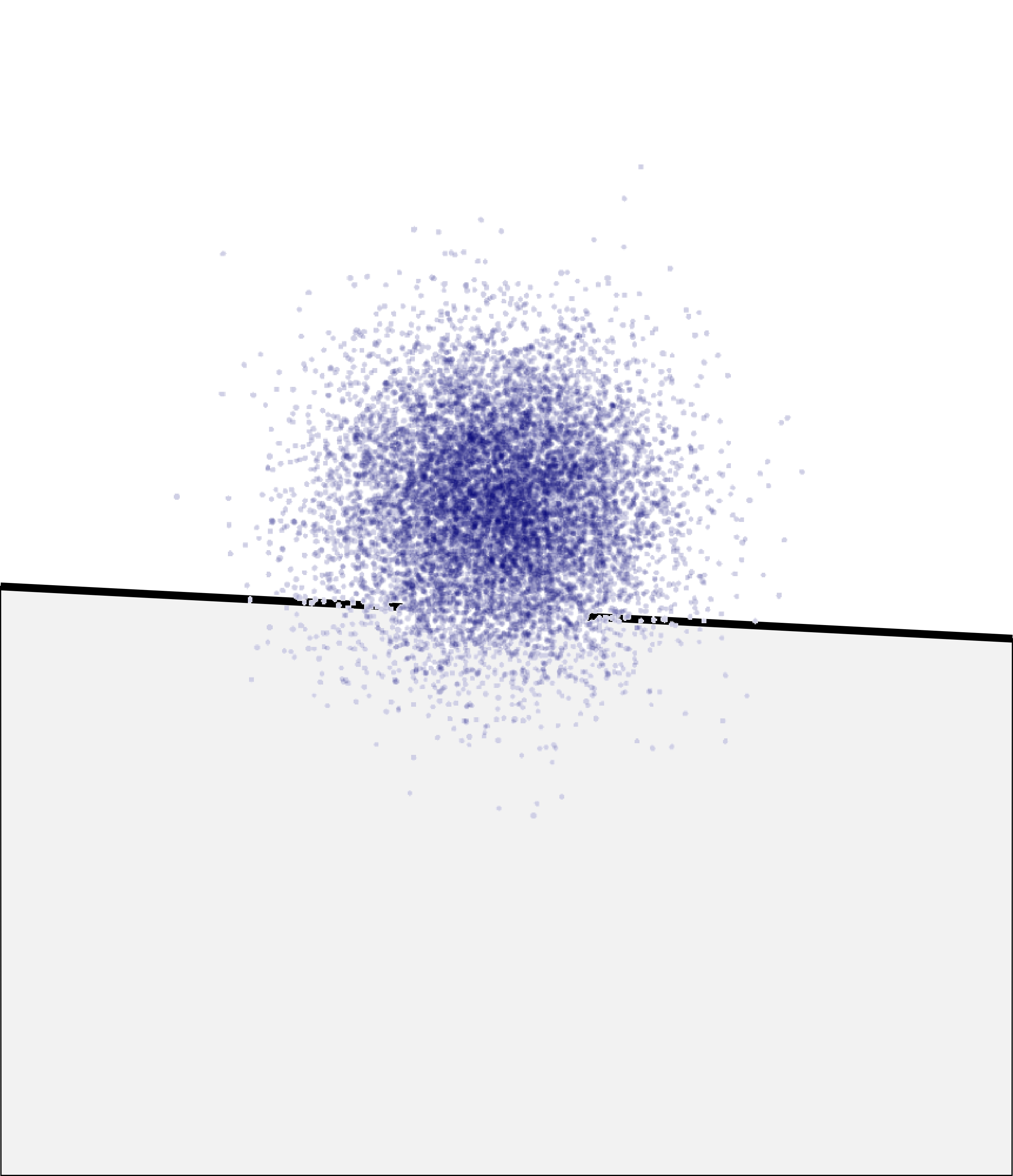}
	\includegraphics[width=0.240\columnwidth]{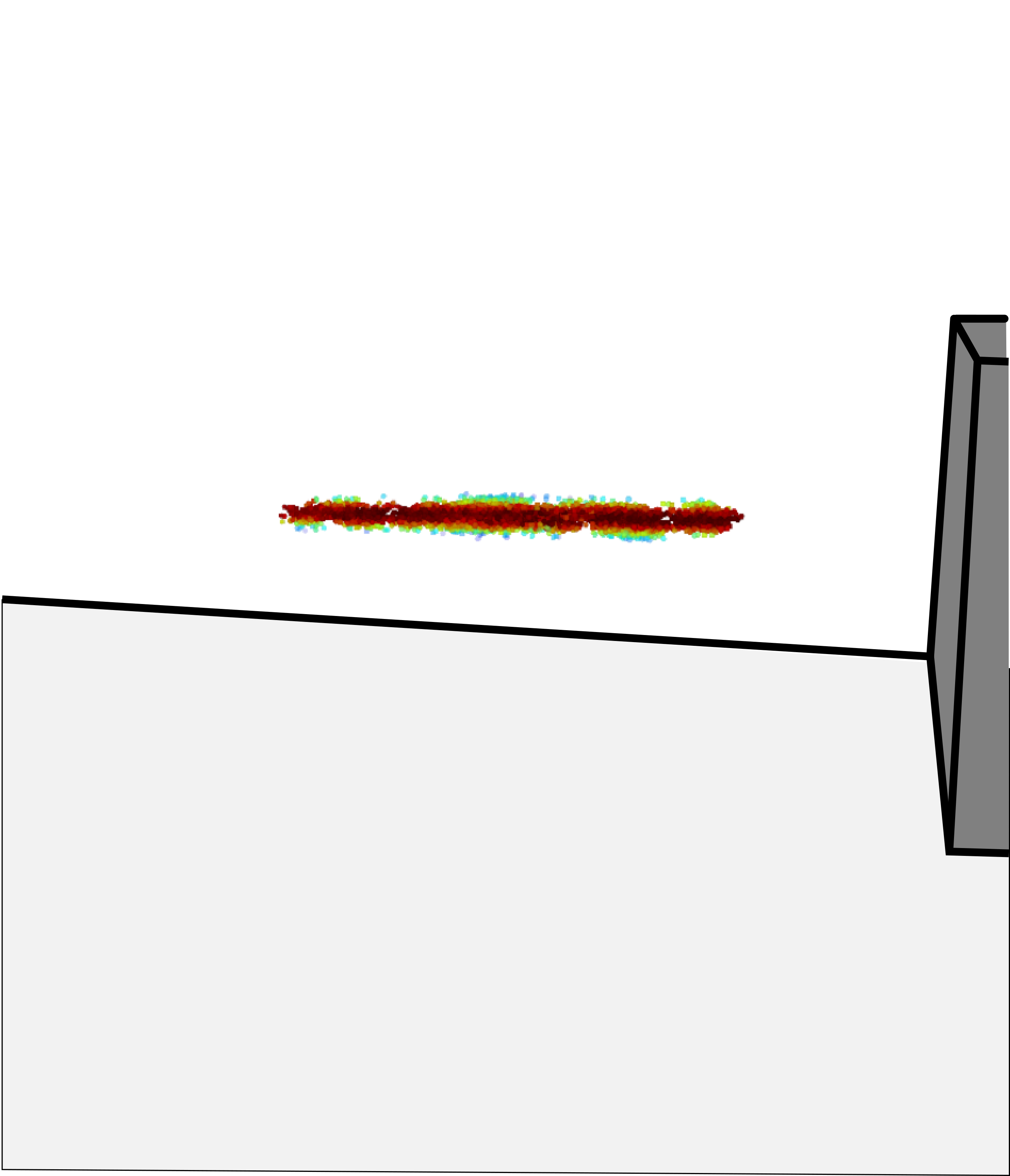}
	\includegraphics[width=0.240\columnwidth]{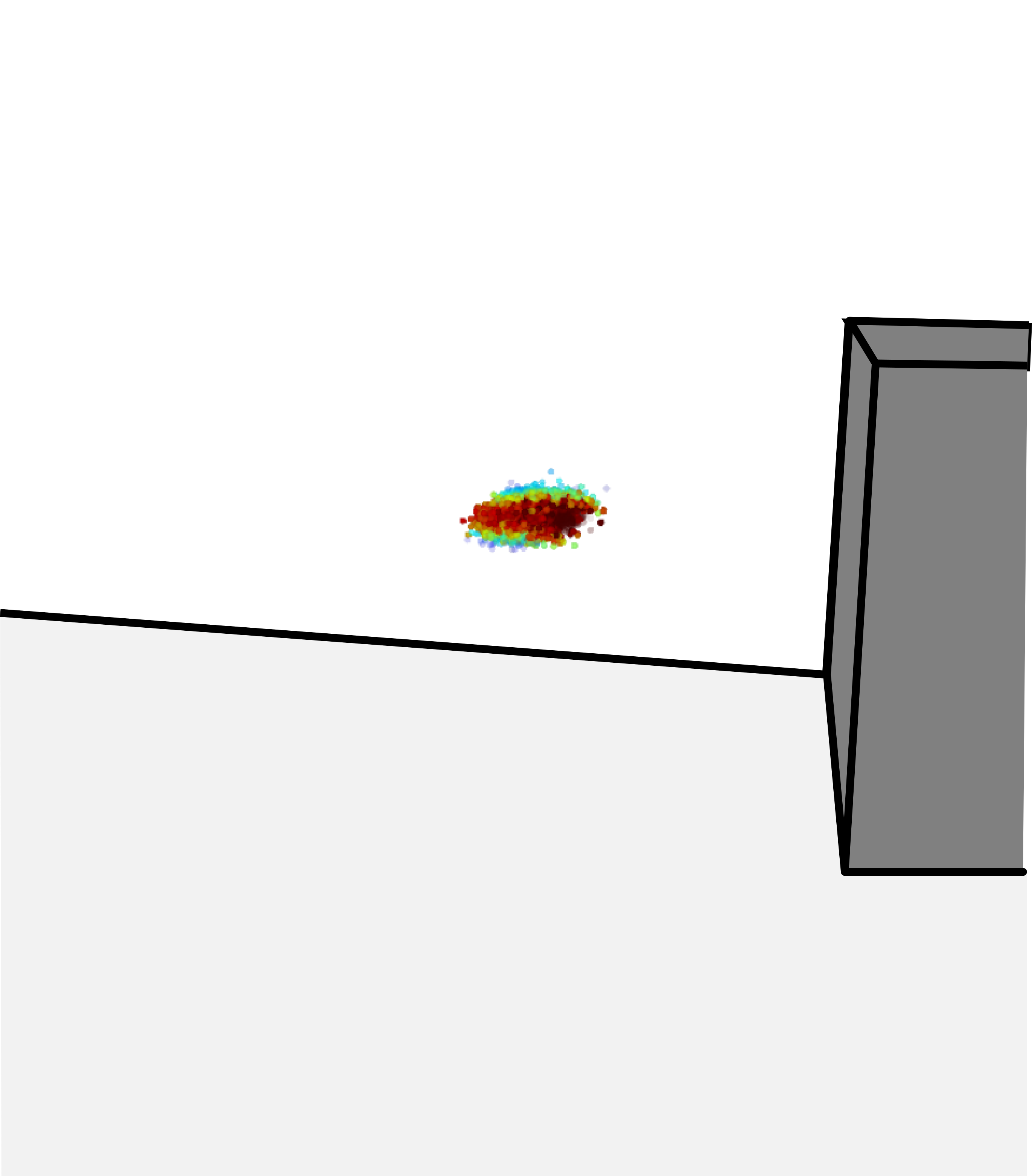}
	\includegraphics[width=0.240\columnwidth]{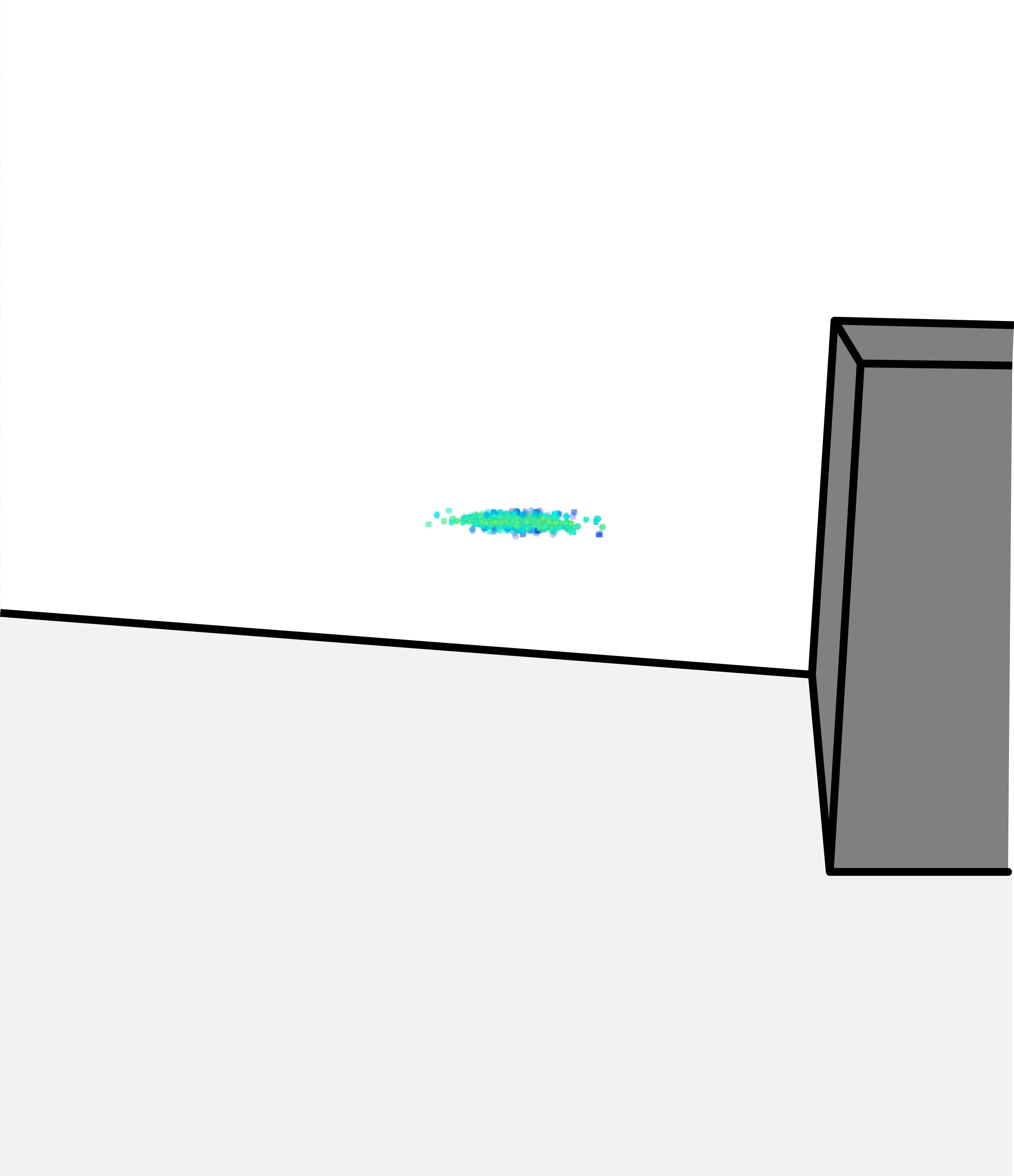}
	\caption{3-dimensional haptic probing experiment. The top row shows the
robot performing the experiment while the bottom row shows the particle
distribution and localization estimate. The
particle set is colorized by normalized weight according to the jet colormap
(i.e., dark blue = lowest weight, dark red = highest weight).  First, on the
bottom left, the robot has an initial distribution with
poses equally weighed. The robot makes a forward probe and then moves to the
right. Now the particles are distributed as an ellipse with high uncertainty to
the left and right of the robot. Then, the robot makes a probe to the right and
touches
an obstacle; the particle cloud collapses into a tight cluster. Since the
robot is now localized, it is able to complete the task of pressing the button
on the wall.}
	\label{fig:3d-probing}
\end{figure}

As shown in Fig.~\ref{fig:3d-probing} the robot was able to correct its
localization and complete the task of touching the button. The initial probe to
the front reduced uncertainty in the robot's $x$ and $z$ directions, which
reduces the particles to an ellipsoidal elongated along $y$. As the robot moves,
uncertainty in the $x$ direction increases slightly. By touching the wall on
the right, the robot re-localized in all three dimensions in much the same way
as a human following a wall in the dark would. The re-localization allows the
robot to press the button, demonstrating the generalization of our algorithm to
3D.
The final position error was: $[7.7, -3.7, -0.2]$ centimeters in the $x$, $y$
and
$z$ directions\footnote{A video showing these experiments is available online: https://youtu.be/VfA8DAyr8t0}.

\section{Discussion}
\label{sec:discussion}
\begin{figure}
 \centering
 \includegraphics[width=0.240\columnwidth]{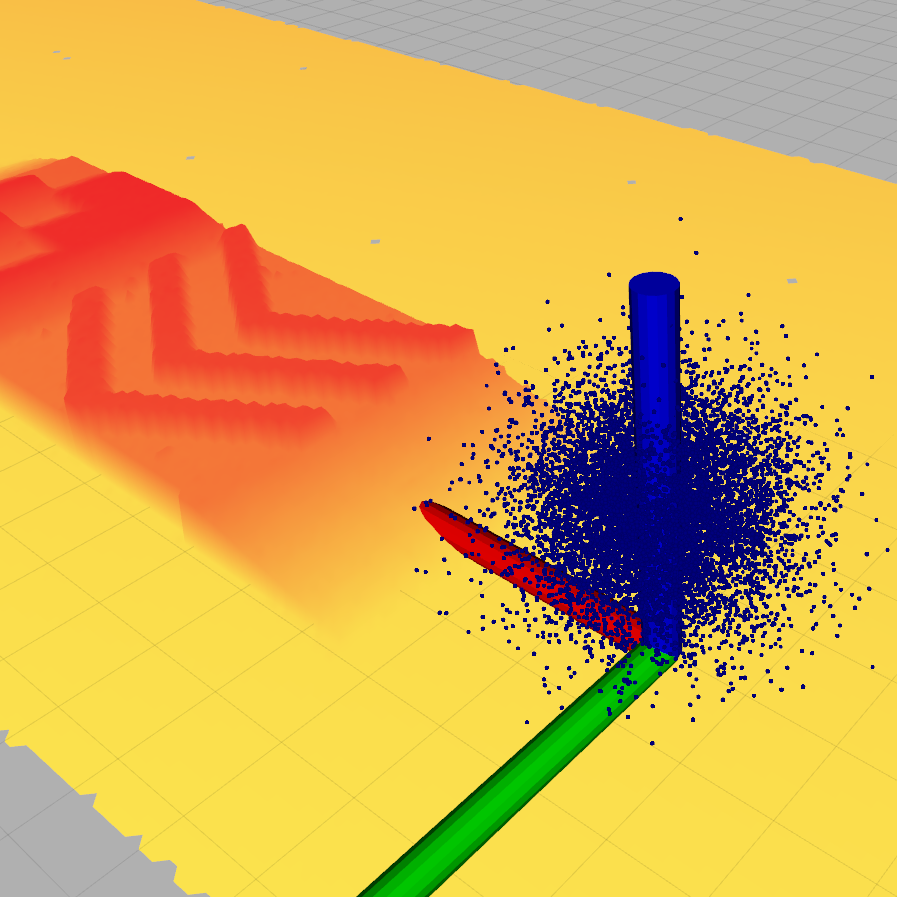}
 \includegraphics[width=0.240\columnwidth]{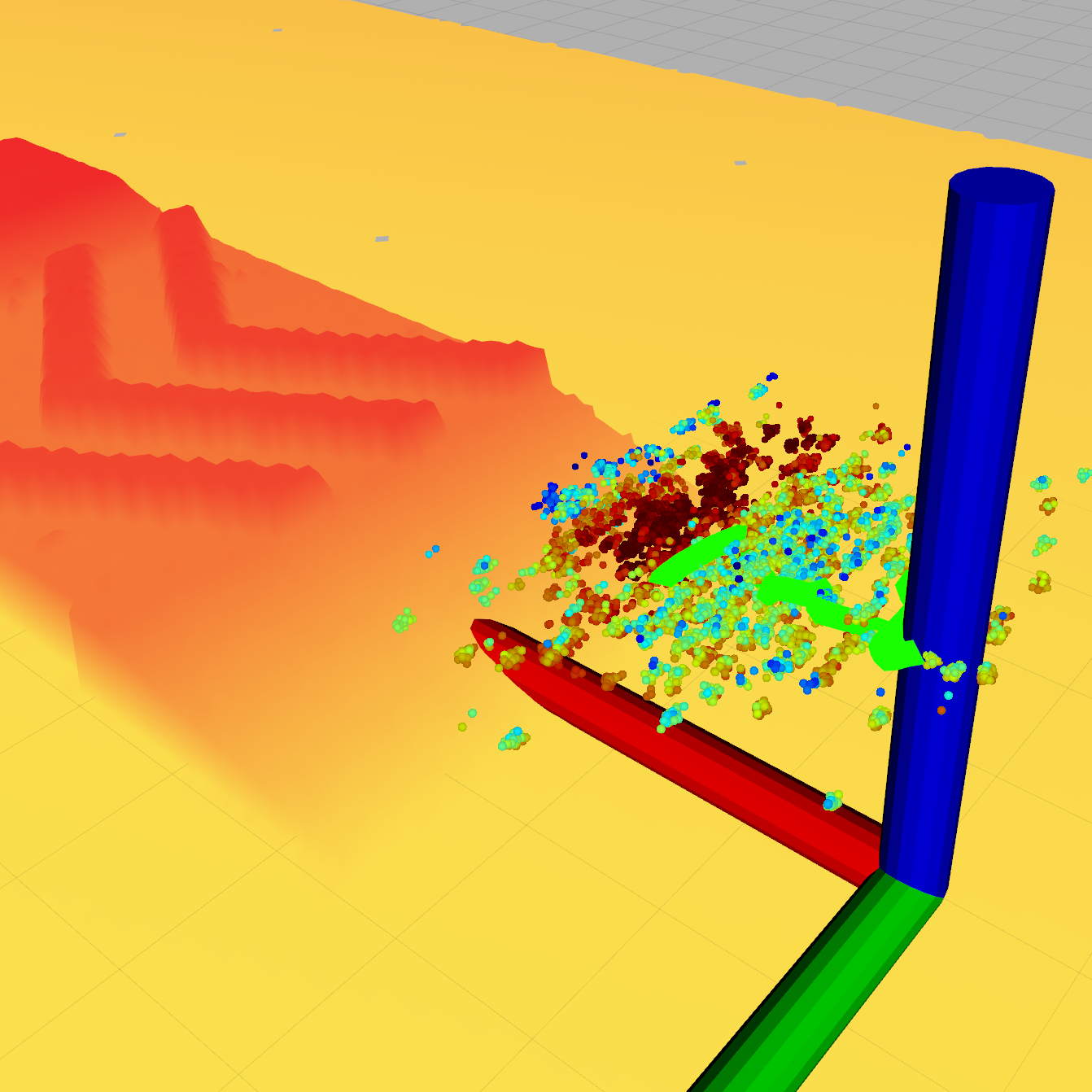}
 \includegraphics[width=0.240\columnwidth]{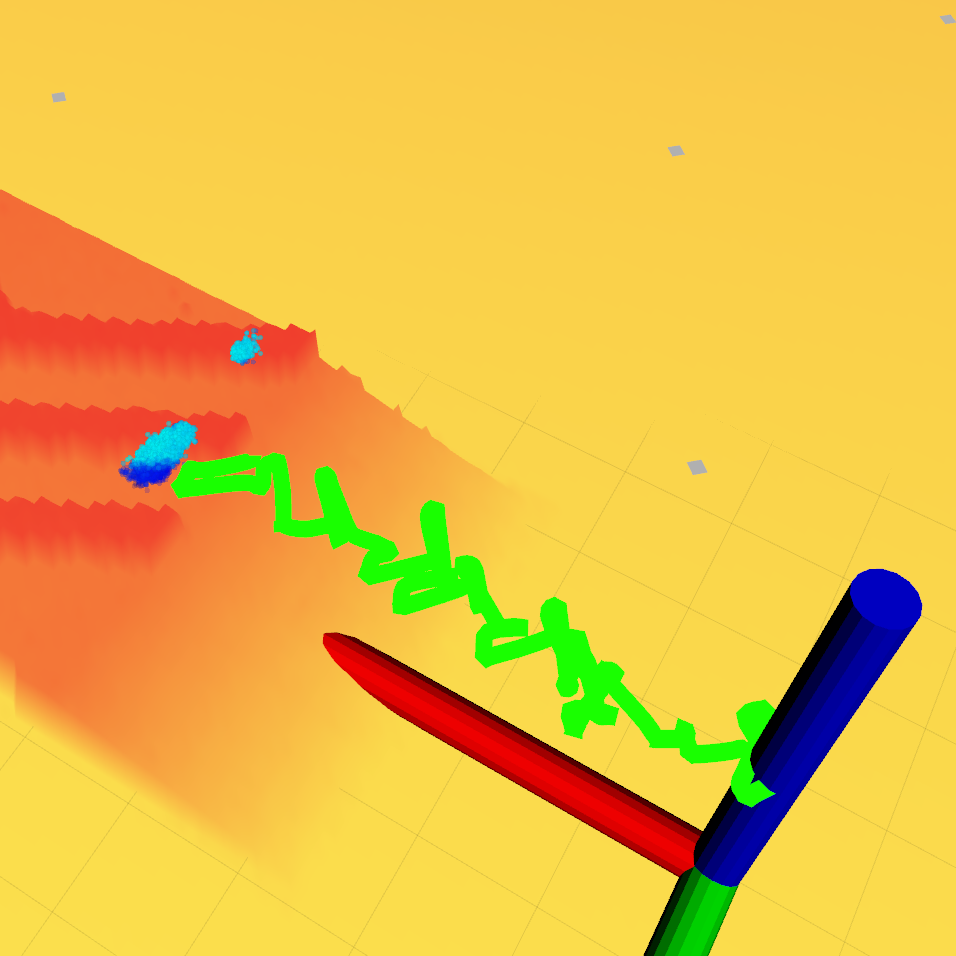}
 \includegraphics[width=0.240\columnwidth]{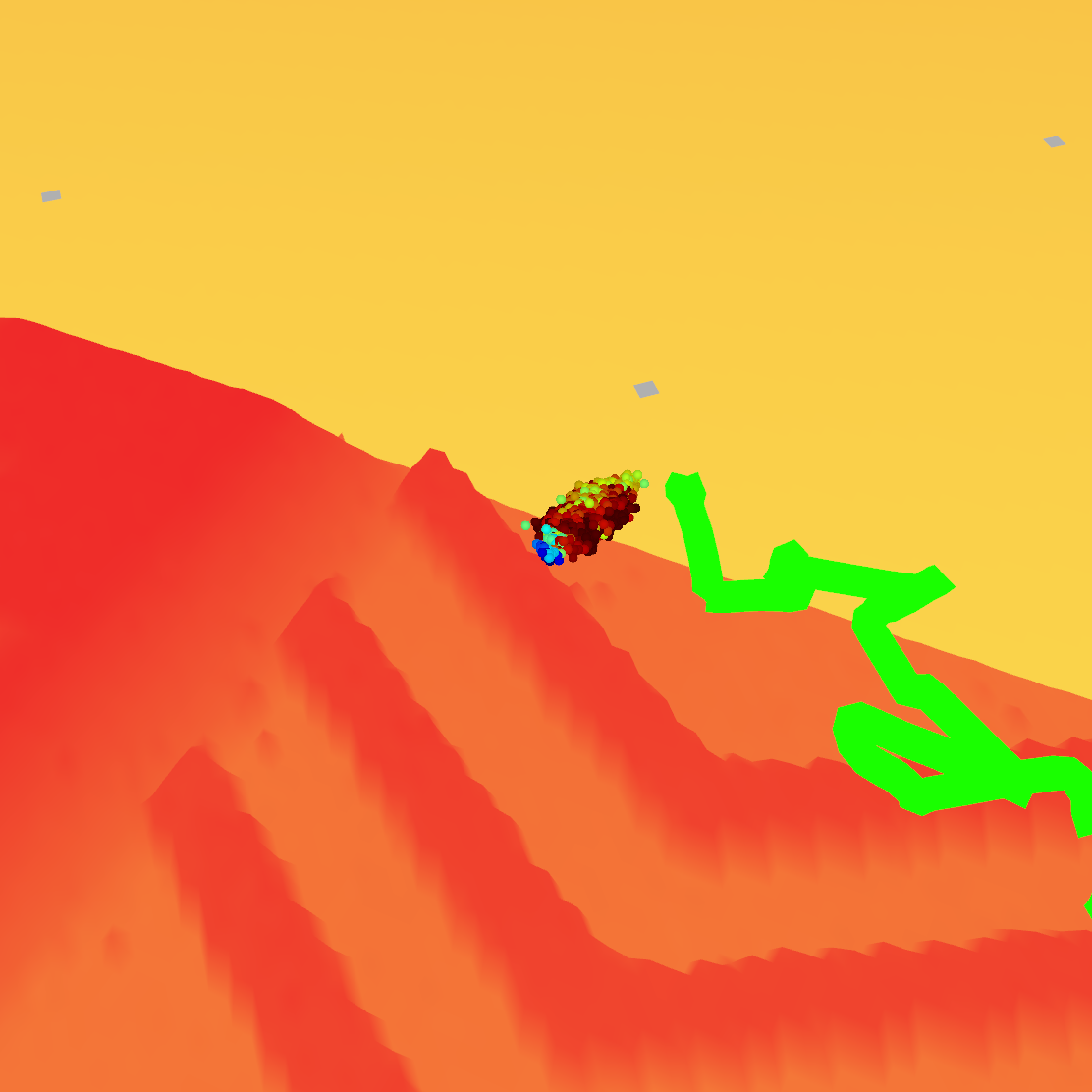}
 \caption{Evolution of the particle set during a trial on the terrain course. The
particle set is colorized by normalized weight according to the jet colormap
(i.e., dark blue = lowest weight, dark red = highest weight). The
green line indicates the ground truth trajectory. \emph{A)} At start,
all the particles have the same weight and are normally distributed at the
starting position. \emph{B)} After a few steps on the ramp, the robot
pose is well estimated on $x$ and $z$ directions, but there is uncertainty on
$y$. \emph{C)} When the robot approaches the chevron the particle set divides
into two clusters, indicating two strong hypotheses as to the robot pose.
\emph{D)} After a few more steps on the chevron, the robot is fully
localized and the particles are tightly clustered. }
 \label{fig:particles}
 \vspace{-10pt}
\end{figure}

The results presented in Section \ref{sec:experimental} demonstrate that a
terrain with moderate complexity such as
the one shown in Fig. \ref{fig:anymal-chevron} already provides enough
information to bound the uncertainty of the robot's location. The effectiveness is obviously limited by the actual terrain morphology in a
real world
 situation, which
would need to contain enough features such that all the DoF of the robot are
constrained once the robot has touched them. Fig.
\ref{fig:particles} shows the evolution of the particles up to the first half
of the terrain course for Experiment 2. As the robot walks through, the
particle swarm gets tighter, indicating good convergence to the most likely
robot pose.

In the third subfigure, it
can be noted how the probability distribution over the robot's pose follows
a bimodal distribution which is visible as two distinct clusters of particles.
This situation justifies the use of
particle filters, as they are able to model non-Gaussian distributions which
can arise from a particular terrain morphology. In this
case, the bimodal distribution is caused by the two identical gaps in between
the chevrons. In such situations, a weighted average of the particle swarm
would lead to a poor approximation of the true pose distribution. Therefore,
the particle analysis described in Section \ref{sec:statistics} is crucial
to reject such an update.

\section{Conclusion}
\label{sec:conclusion}
We have presented a haptic localization algorithm for quadruped robots based on
Sequential Monte Carlo methods. The algorithm can reduce the uncertainty of the
estimated robot's trajectory walking over a non-degenerate terrain course, with
an average pose estimation accuracy up to \SI{10}{\centi\meter}. This is
fundamental for repetitive autonomous tasks in
vision-denied conditions such as inspections in sewage systems. With routine
inspection tasks in mind, we have also implemented a haptic exploration motion
to localize against walls. In future work, we are planning to extend the
localization algorithm to incorporate more features into the measurement model,
such as estimated slope and friction, to further improve the localization
performance as well as active planning for direct the probing actions.

\section{Acknowledgments}
This research has been conducted as part of the ANYbotics research community. It
was part funded by the EU H2020 Project THING and a Royal Society University
Research Fellowship (Fallon).

\bibliographystyle{./IEEEtran}
\bibliography{./IEEEabrv,./library}
\end{document}